\documentclass[]{xiaomi}

\usepackage[utf8]{inputenc} 
\usepackage[T1]{fontenc}    
\usepackage{hyperref}       
\usepackage{url}            
\usepackage{booktabs}       
\usepackage{amsfonts}       
\usepackage{nicefrac}       
\usepackage{microtype}      
\usepackage{xcolor}         

\usepackage{overpic}
\usepackage{rotating}   
\usepackage{graphicx}   
\usepackage{array}
\usepackage[table]{xcolor}
\usepackage{pifont}
\usepackage{textcomp}
\usepackage{multirow} 
\usepackage{subcaption}   
\usepackage{amsmath}
\usepackage{marvosym}   
\usepackage{ragged2e}   
\usepackage{enumitem}

\usepackage{xspace}

\usepackage{wrapfig}  

\title{ParkGaussian: Surround-view 3D Gaussian Splatting for Autonomous Parking}

\author[\textasteriskcentered 1]{Xiaobao Wei}
\author[\textasteriskcentered 2]{Zhangjie Ye}
\author[\textasteriskcentered 2]{Yuxiang Gu}
\author[2]{Zunjie Zhu}
\author[2]{Yunfei Guo}
\author[1]{Yingying Shen}
\author[1]{Shan Zhao}
\author[2]{Ming Lu}
\author[\dagger 1]{Haiyang Sun}
\author[1]{Bing Wang}
\author[1]{Guang Chen}

\author[\textsuperscript{\Letter} 2]{Rongfeng Lu}
\author[\textsuperscript{\Letter} 1]{Hangjun Ye}

\affiliation[1]{Xiaomi EV}
\affiliation[2]{Hangzhou Dianzi University}

\contribution[\textasteriskcentered]{Equal Contribution}
\contribution[\dagger]{Project Leader}
\contribution[\textsuperscript{\Letter}]{Corresponding Author}

\abstract{
Parking is a critical task for autonomous driving systems (ADS), with unique challenges in crowded parking slots and GPS-denied environments. However, existing works focus on 2D parking slot perception, mapping, and localization, 3D reconstruction remains underexplored, which is crucial for capturing complex spatial geometry in parking scenarios. 
Naively improving the visual quality of reconstructed parking scenes does not directly benefit autonomous parking, as the key entry point for parking is the slots perception module. 
To address these limitations, we curate the first benchmark named ParkRecon3D,  specifically designed for parking scene reconstruction. It includes sensor data from four surround-view fisheye cameras with calibrated extrinsics and dense parking slot annotations. We then propose ParkGaussian, the first framework that integrates 3D Gaussian Splatting (3DGS) for parking scene reconstruction. 
To further improve the alignment between reconstruction and downstream parking slot detection, we introduce a slot-aware reconstruction strategy that leverages existing parking perception methods to enhance the synthesis quality of slot regions. 
Experiments on ParkRecon3D demonstrate that ParkGaussian achieves state-of-the-art reconstruction quality and better preserves perception consistency for downstream tasks. The code and dataset will be released. 
}

\date{\today} 
\project{https://wm-research.github.io/ParkGaussian/}

\begin{document}
\thispagestyle{firstheader}
\maketitle

\section{Introduction}

\label{sec:intro} 

Autonomous parking is a vital component of autonomous driving systems (ADS)~\cite{olmos2025overview}. Unlike on-road driving, which typically occurs in structured and GPS-available environments, parking often takes place in narrow underground areas, crowded parking slots, and poor illumination~\cite{millard2019autonomous}. These factors pose challenges for accurate perception and localization, highlighting the necessity of specialized techniques for parking scenarios. 

Early works~\cite{huang2019dmpr, min2021attentional_gcn, zinelli2019deep, li2020vacant, wu2020psdet, lu2023open, yu2022mttrans} concentrate on parking perception, particularly parking slot detection, to identify and localize slots in surround-view images. These methods typically employ Inverse Perspective Mapping (IPM) to transform multi-view fisheye images into a Bird’s-Eye View (BEV) representation for slot perception. 
Building upon these detection modules, subsequent studies~\cite{tripathi2020trained, li2024avm, AVPSLAM2020} investigate Simultaneous Localization and Mapping (SLAM) in parking scenes, where parking slot landmarks serve as reliable references to enhance localization robustness in GPS-denied and visually repetitive environments. 
Recently, several approaches~\cite{rathour2018vision, hu2022st, yang2024e2e, chen2025end} explore end-to-end frameworks that jointly learn perception, planning, and control from sensor inputs, training on simulated parking datasets collected in CARLA~\cite{2017_11_10-CARLA-An_Open_Urban_Driving_Simulator}. 

Although effective for simulated parking planning, there remains a large gap between CARLA and real-world parking environments. This gap highlights the need for a realistic simulator to assess slot perception and closed-loop planning performance in complex parking scenes. 

To bridge the gap between simulators and real-world environments, recent research~\cite{2017_11_10-CARLA-An_Open_Urban_Driving_Simulator, 2017_7_18-AirSim-High_Fidelity_Visual_and_Physical_Simulation_for_Autonomous_Vehicles, dauner2024navsim} has increasingly focused on realistic driving simulation through 4D scene modeling, encompassing both reconstruction and generation approaches for street-level scenarios. For reconstruction, early methods~\cite{2023_3_25-SUDS-Scalable_Urban_Dynamic_Scenes, 2021_3_5-Neural_Scene_Graphs_for_Dynamic_Scenes, 2022_2_10-Block-NeRF-Scalable_Large_Scene_Neural_View_Synthesis, 2023_11_26-NeuRAD-Neural_Rendering_for_Autonomous_Driving, 2023_7_27-MARS-An_Instance_aware_Modular_and_Realistic_Simulator_for_Autonomous_Driving, 2023_11_3-EMERNERF-EMERGENT_SPATIAL_TEMPORAL_SCENE_DECOMPOSITION_VIA_SELF-SUPERVISION} build upon Neural Radiance Fields (NeRF)~\cite{2020_08_03-NeRF-Representing_Scenes_as_Neural_Radiance_Fields_for_View_Synthesis, wei2024nto3d} to represent street scenes, marking the initial exploration for driving environments. But these approaches suffer from low efficiency. Recent advances~\cite{chen2025mixedgaussianavatar, wang2025plgs, wei2025gazegaussian, wei2025graphavatar} based on 3D Gaussian Splatting (3DGS)~\cite{2023_8_08-3dgs_for_real_time_radiance_field_rendering, qiwu3dgut, Liao2024FisheyeGS} have significantly improved rendering speed, which represents dynamic street scenes in a box supervised~\cite{2024_01_02-street_gaussian-modelling_dynamic_urban_scenes_with_gs, chen2024omnire, 2024_2_27-drivinggaussian-compusing_gaussian_splatting_for_surronding_dynamic_autonomous_driving_scenes, 2024_3_19-HUGS-Holistic_Urban_3D_Scene_Understanding_via_Gaussian_Splatting, hess2025splatad} or a self-supervised manner~\cite{2024_5_30-S3Gaussian-Self_Supervised_Street_Gaussians_for_Autonomous_Driving, 2024_6_26-VDG-Vision-Only_Dynamic_Gaussian_for_Driving_Simulation, 2024_3_20-PVG-periodic_vibration_gaussian-dynamic_urban_scene_reconstruction, peng2025desire, wei2025emd}. For generation, recent models based on diffusion and controllable generation frameworks~\cite{LDM, gao2023magicdrive, guo2025genesis, zeng2025rethinking, li2025manipdreamer3d, li2025manipdreamer} synthesize street scenes from scene layouts or text. 

\begin{wrapfigure}{r}{0.5\textwidth} 
  \centering
  \vspace{-10pt} 
  \includegraphics[width=\linewidth]{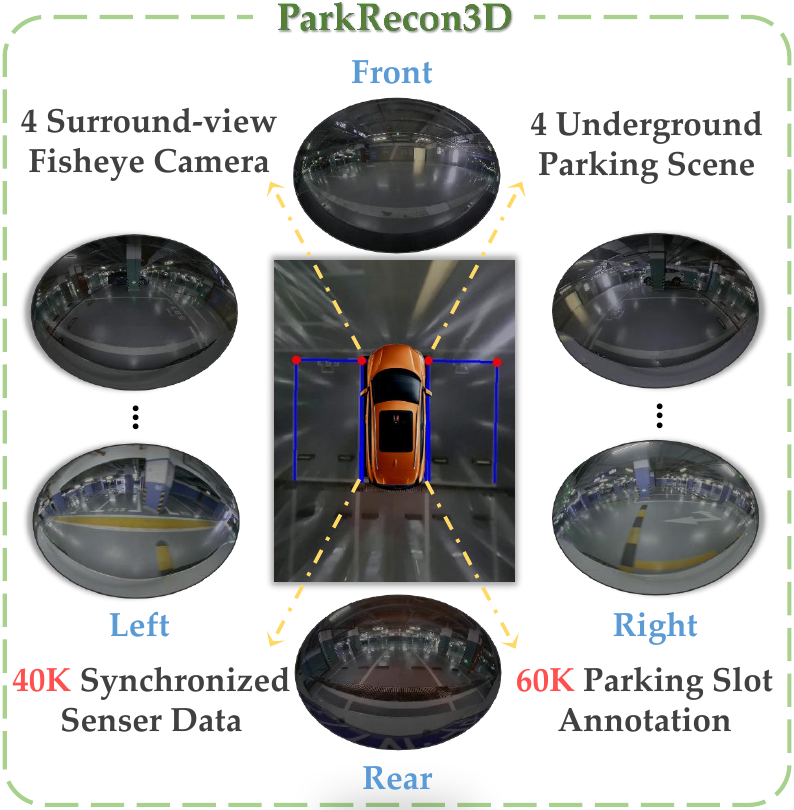} 
  \caption{\textbf{ParkRecon3D}: a surround-view dataset for parking-aware reconstruction. 
    It contains synchronized image streams from four fisheye cameras covering full 360° surroundings and four representative underground parking scenes. 
    In total, the dataset includes 40K multi-camera sensor frames and 60K human-annotated parking-slot labels, supporting parking-slot perception and geometry-aware 3D reconstruction.}
  \label{fig:teaser}
  \vspace{-10pt} 
\end{wrapfigure}

These methods achieve photorealistic synthesis quality in typical on-road driving scenes. Building upon these, several works further develop reinforcement learning pipelines~\cite{gao2025rad} and closed-loop simulations~\cite{yan2025drivingsphere, jiang2025realengine, yang2025drivearena} for training and evaluating autonomous driving systems~\cite{cao2025fastdrivevla, liao2025diffusiondrive}. 

However, existing driving simulators primarily focus on modeling on-road driving, leaving parking scene simulation largely unexplored. 
Additionally, previous reconstruction methods mainly emphasize visual fidelity while overlooking the fundamental goal of simulation, which is to generate perception-aligned synthetic data for faithfully evaluating the performance and limitations of downstream models. 
For autonomous parking, where the parking slot detection module serves as the system’s entry, it is crucial to align visual fidelity in slot-related regions with the downstream perception models. 


To address these limitations, we first curate a benchmark named \textbf{ParkRecon3D} (Fig.~\ref{fig:teaser}), specifically designed for parking-scene reconstruction. ParkRecon3D is built upon the open-sourced dataset from AVM-SLAM~\cite{AVMSLAM2024}, containing sensor data captured by four calibrated fisheye cameras in underground parking lots. ParkRecon3D provides over 40K synchronized sensor frames and 60K annotated parking slots with accurate extrinsic calibration. Building on this benchmark, we propose \textbf{ParkGaussian}, the first framework that integrates 3D Gaussian Splatting (3DGS) adapted to surround-view fisheye images to achieve high-quality 3D reconstruction of parking scenes.  Furthermore, we introduce a slot-aware reconstruction strategy that leverages two widely used parking-slot detection methods (DMPR-PS~\cite{huang2019dmpr} and GCN-Parking~\cite{min2021attentional_gcn}) to identify slot regions in a learnable manner. This strategy not only improves synthesis quality in slot areas but also enhances alignment with downstream perception tasks, establishing a reliable simulator for autonomous parking systems.  

Our main contributions are summarized as follows: 
\begin{itemize}
\item We curate \textbf{ParkRecon3D}, the first benchmark tailored for 3D reconstruction in parking scenarios, featuring over 40K synchronized fisheye sensor frames and 60K precisely annotated parking slots with calibrated extrinsics, captured in underground parking lots. 
\item We present \textbf{ParkGaussian}, a novel framework that adapts 3DGS to surround-view fisheye sensors and incorporates a slot-aware reconstruction strategy leveraging parking slot detectors to enhance reconstruction fidelity in task-critical slot regions. 
\item We conduct extensive experiments on ParkRecon3D, demonstrating that ParkGaussian achieves state-of-the-art reconstruction quality and stronger alignment with downstream perception. 
\end{itemize}

\section{Related Work}
\label{sec:related_works}

\paragraph{Driving Scene Simulation.}
Traditional autonomous driving simulators~\cite{2017_7_18-AirSim-High_Fidelity_Visual_and_Physical_Simulation_for_Autonomous_Vehicles, 2017_11_10-CARLA-An_Open_Urban_Driving_Simulator} demand substantial manual effort to construct environments and fall short of achieving photorealistic rendering. 
The emergence of neural field based approaches~\cite{2021_3_5-Neural_Scene_Graphs_for_Dynamic_Scenes, 2023_7_27-MARS-An_Instance_aware_Modular_and_Realistic_Simulator_for_Autonomous_Driving, 2023_11_26-NeuRAD-Neural_Rendering_for_Autonomous_Driving, 2024_3_29-Multi-Level_Neural_Scene_Graphs_for_Dynamic_Urban_Environments} has enabled realistic rendering and achieved a scalable neural scene representation for large-scale urban environments. 
Subsequently, 3DGS-based methods decompose street scenes in a box supervised~\cite{2024_01_02-street_gaussian-modelling_dynamic_urban_scenes_with_gs, chen2024omnire, 2024_2_27-drivinggaussian-compusing_gaussian_splatting_for_surronding_dynamic_autonomous_driving_scenes, 2024_3_19-HUGS-Holistic_Urban_3D_Scene_Understanding_via_Gaussian_Splatting, hess2025splatad} or a self-supervised manner~\cite{2024_5_30-S3Gaussian-Self_Supervised_Street_Gaussians_for_Autonomous_Driving, 2024_6_26-VDG-Vision-Only_Dynamic_Gaussian_for_Driving_Simulation, 2024_3_20-PVG-periodic_vibration_gaussian-dynamic_urban_scene_reconstruction, peng2025desire, wei2025emd}. 
For example, OmniRe~\cite{chen2024omnire} constructs hierarchical scene representations that integrate dynamic object graphs, unifying the modeling of static backgrounds and dynamic entities such as vehicles and pedestrians to achieve high-fidelity reconstruction of complex urban driving environments. 

However, these methods primarily focus on driving scenes and are not suitable for underground parking environments. To achieve promising synthesis quality, they heavily rely on accurate multi-sensor setups, including dense LiDAR point clouds and calibrated GPS/IMU localization. In contrast, underground parking scenes present a much harsher setting, characterized by poor illumination, limited texture diversity and the absence of GPS signals required for accurate extrinsic calibration. These factors pose significant challenges for parking scene reconstruction.

\paragraph{Autonomous Parking.} 
Autonomous parking serves as a foundation of ADS. Compared to driving, it presents more challenges in perception and localization, as parking lots are characterized as narrow and GPS-denied. 
Previous works in parking slot detection~\cite{huang2019dmpr, min2021attentional_gcn} simplify this task by translating surround-view images to BEV representation, which inevitably limits the ability to perceive the entire space. 
AVP-SLAM~\cite{AVPSLAM2020} and AVM-SLAM~\cite{AVMSLAM2024} provide a method for automatically navigating into parking lots and parking in the target spot by extracting semantic features from stable parking slots, highlighting the importance of parking slot detection. 
Beyond SLAM, there has been a growing trend toward end‑to‑end models~\cite{rathour2018vision, hu2022st, yang2024e2e, chen2025end} that integrate perception, planning, and control within a unified learning framework for autonomous parking in the simulation environment~\cite{2017_11_10-CARLA-An_Open_Urban_Driving_Simulator}. Despite the appeal of fully end-to-end pipelines, the entry point of autonomous parking remains perception, and the quality of perception directly determines downstream planning accuracy. 
In particular, parking slot detection relies on recognizing fine-grained, geometry-driven patterns such as L-shaped and T-shaped corners, making local geometric consistency crucial. While previous studies have advanced parking specialized methods, research dedicated to 3D parking reconstruction remains scarce. 



To bridge this gap, we introduce ParkRecon3D, the first benchmark specifically tailored for 3D reconstruction in parking scenarios, and ParkGaussian, a novel 3D Gaussian Splatting framework adapted to surround‑view fisheye inputs with a slot‑aware reconstruction strategy. 
\begin{figure*}[!ht] 
  \centering
  \includegraphics[width=0.95\textwidth]{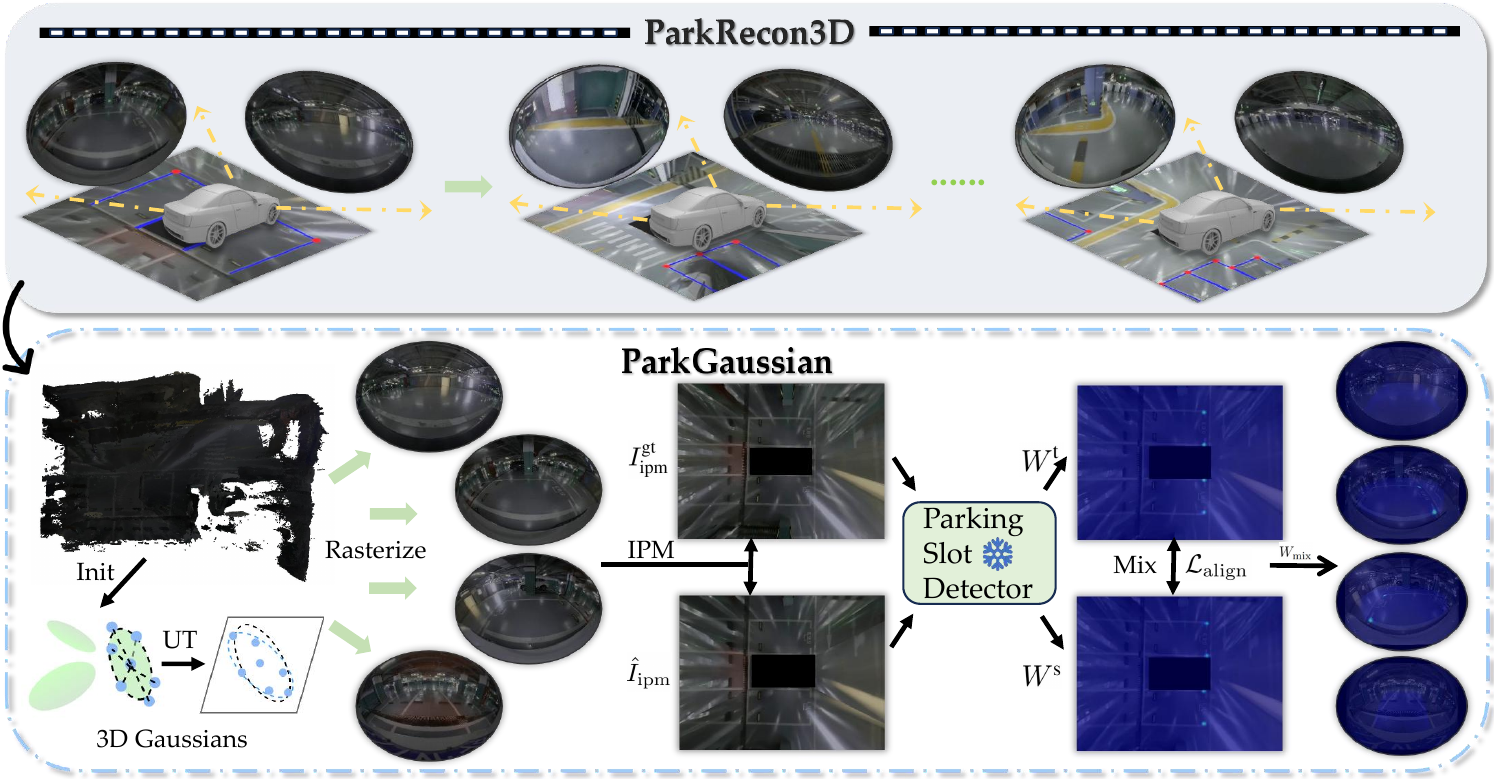} 
  \caption{Overview of the ParkGaussian pipeline. Four surround-view fisheye images are first rendered from 3D Gaussian primitives using UT-based projection for stable splatting under strong distortion. The rendered views are then passed through a differentiable IPM module to obtain a unified BEV map. A pretrained parking-slot detector (DMPR-PS or GCN-Parking) processes both rendered and ground-truth BEV maps to produce teacher–student structural guidance, from which slot-aware weights are constructed. These weights supervise reconstruction in both the IPM space and the camera-view space, forming the alignment and slot-aware objectives. ParkGaussian jointly optimizes Gaussian attributes toward photometric fidelity and perception-aligned slot geometry.}
  \label{fig:pipeline.png}
  \vspace{-4mm}
\end{figure*}

\section{Method}
In this section, we first describe the construction of the ParkRecon3D benchmark. We then review the preliminaries of 3D Gaussian Splatting (3DGS) and its extension with 3DGUT. Finally, we present the IPM projection and the slot-aware reconstruction modules that form the core of ParkGaussian. 

\subsection{ParkRecon3D Benchmark} 
To the best of our knowledge, there is currently no benchmark specifically designed for 3D reconstruction in parking environments. We therefore construct ParkRecon3D by reorganizing and extending the open-source dataset used for AVM-SLAM~\cite{AVMSLAM2024}. The data are collected in an underground parking facility of approximately $220\,\text{m} \times 110\,\text{m}$ with more than 430 parking spaces. The ego vehicle is equipped with a production around-view monitoring system consisting of four fisheye cameras mounted at the front, rear, left, and right sides of the vehicle. Each camera records images at 10\,Hz with a resolution of $1280 \times 960$, and the system additionally produces synthesized Inverse Perspective Map (IPM) images at $1354 \times 1632$ from the four-view fisheye inputs. The sequences are organized into four representative scenes covering diverse parking layouts. 

To provide an accurate geometric reference for reconstruction, we calibrate the extrinsic parameters of all four fisheye cameras using COLMAP~\cite{schoenberger2016sfm,schoenberger2016mvs}, instead of relying on wheel-encoder and IMU odometry that is noisy in underground environments. For parking-slot perception, we follow the annotation protocol of DMPR-PS~\cite{huang2019dmpr} and manually label slot corner points in the BEV domain, yielding high-quality supervision for slot-level detection. Building on this data, we curate ParkRecon3D, the first benchmark tailored for 3D reconstruction in parking scenarios. It contains over 40K synchronized multi-fisheye frames and 60K manually verified parking-slot annotations with calibrated extrinsics. The benchmark will be released to the community as a comprehensive dataset for training and evaluating 3D reconstruction models in underground parking environments.

\subsection{Preliminary}
\paragraph{3D Gaussian Splatting.} 3DGS~\cite{2023_8_08-3dgs_for_real_time_radiance_field_rendering} is an advanced method for representing 3D scenes, using a set of 3D Gaussian spheres to represent the scene. Its probability density function is formulated as:
\begin{equation}
G(x)=e^{-\frac{1}{2}(x-\mu )^{T}\Sigma^{-1}(x-\mu )} 
\end{equation}
where \( \mu\in\mathbb{R}^3 \) is the mean value, which is the coordinates of the 3D Gaussian sphere in the world; \( \Sigma \) represents the covariance matrix, which can be decomposed into \( \Sigma = RSS^{T}R^{T} \), where \( {R}\in SO(3)\) is the rotation matrix and \( {T} \in \mathbb{R}^{3 \times 3}\) is the scaling matrix, which is stored by the rotation quaternion \( {q} \in \mathbb{R}^4\). For the colors in the pixels, the spherical harmonics (SH) and the opacity \( \alpha \) are assigned to the color of the Gaussian sphere, and the depth sorting is carried out according to the distance between the Gaussian sphere and the camera, and then the color of a pixel can be calculated by using \( \alpha \) - blended:
\begin{equation}
\mathbf{c} = \sum_{i=1}^{N} \mathbf{c}_i \alpha_i \prod_{j=1}^{i-1} (1 - \alpha_j)
\end{equation}

\paragraph{3DGS with UT-based projection.}
Vanilla 3DGS relies on an EWA splatting formulation that linearizes the camera projection using a first-order Jacobian. This becomes inaccurate for heavily distorted fisheye cameras and requires camera-specific Jacobian derivations, which is undesirable for multi-camera surround-view systems. To address this issue, we incorporate the unscented-transform (UT) projection used in 3DGUT~\cite{qiwu3dgut} as a drop-in replacement within 3DGS.

Instead of linearizing the non-linear projection $v = g(x)$, the UT approximates the 3D Gaussian $\mathcal{N}(\mu, \Sigma)$ using a small set of sigma points: 
\begin{equation}
x_i = 
\begin{cases}
\mu, & i = 0,\\
\mu \pm \sqrt{(n+\lambda)\Sigma}_{[i]}, & i = 1,\dots,2n,
\end{cases}
\label{eq:sigma_points}
\end{equation}
which are projected exactly through $g(\cdot)$.
The mean and covariance of the resulting 2D Gaussian footprint are then computed as: 
\begin{equation}
\begin{aligned}
\mu_v &= \sum_i w_i^\mu\, g(x_i), \\
\Sigma_v &= \sum_i w_i^\Sigma\, (g(x_i)-\mu_v)(g(x_i)-\mu_v)^\top
\end{aligned}
\label{eq:ut_mean_var}
\end{equation}

This UT-based projection removes the need for Jacobian derivation for each fisheye camera model and yields more stable 2D Gaussian footprints under strong distortion. In ParkGaussian, this module allows 3DGS to train directly on surround-view fisheye imagery and significantly improves geometric stability in underground parking scenarios. 

\subsection{Differentiable Surround-View IPM}
Most parking slot detection models operate on bird’s-eye-view (BEV) images rather than raw fisheye inputs. Therefore, directly rendering reconstructed fisheye views from 3DGS cannot be used by downstream detectors.  
Thus, we convert the rendered surround-view fisheye images into a unified BEV representation through a fully differentiable Inverse Perspective Mapping (IPM) module. 

Following AVP-SLAM~\cite{AVPSLAM2020}, each surround-view camera is calibrated offline with intrinsic matrix $K_c$ and extrinsic parameters $(R_c, t_c)$.  
A fisheye pixel ${\mathbf{u}} = [u, v, 1]^\top$ is first unprojected to a ray in the camera frame using the inverse fisheye model $\pi_c^{-1}(\cdot)$, and then intersected with the ground plane ($z=0$) in the vehicle frame:
\begin{equation}
\frac{1}{\lambda}
\begin{bmatrix}
x_v \\[3pt]
y_v \\[3pt]
1
\end{bmatrix}
=
\left[R_c \;\; t_c \right]^{-1}_{\text{col}:1,2,4}\,
\pi_c^{-1}
\!\left(
\begin{bmatrix}
u \\[3pt]
v \\[3pt]
1
\end{bmatrix}
\right)
\label{eq:ground_plane_ipm}
\end{equation}
where $[x_v, y_v]^\top$ is the ground-plane point in the vehicle coordinate system, and $\lambda$ is the scale factor to enforce $z=0$.

Then ground-plane points from all four fisheye cameras are fused onto a canonical BEV plane. Given the intrinsic matrix $K_{\text{ipm}}$ of the synthesized IPM image, each point is reprojected to BEV pixel coordinates:
\begin{equation}
\begin{bmatrix}
u_{\text{ipm}} \\[3pt]
v_{\text{ipm}} \\[3pt]
1
\end{bmatrix}
=
K_{\text{ipm}}
\begin{bmatrix}
x_v \\[3pt]
y_v \\[3pt]
1
\end{bmatrix}
\label{eq:ipm_reprojection}
\end{equation}
producing a unified BEV map $\hat{I}_{\text{ipm}}$ consistent with the geometric assumptions of existing parking-slot detectors. All steps in the IPM are expressed in closed-form and implemented as differentiable operations.  
This enables gradients from the downstream detection models to be propagated back to the 3D Gaussian representation, allowing the reconstruction to be directly optimized toward downstream parking slot perception objectives.

\subsection{Slot-aware Reconstruction}
Based on the differentiable IPM mapping, we further introduce Slot-Aware Reconstruction, which injects task-driven supervisory signals from pretrained parking slot detectors directly into the 3DGS optimization process. This encourages Gaussian primitives to encode slot-critical geometry and results in reconstructions that are not only photometrically accurate but also structurally faithful for downstream perception.

After rendering the surround-view fisheye images $\hat{I}_\text{sur}$ from 3DGS, the differentiable IPM operator $\Phi_{\text{IPM}}$ produces a synthesized IPM image
\begin{equation}
\hat{I}_{\text{ipm}} = \Phi_{\text{IPM}}(\hat{I}_\text{sur})
\end{equation}
In parallel, we construct a ground-truth IPM image $I_{\text{ipm}}^{\text{gt}}$ from the annotated ParkRecon3D data using the same IPM mapping. A pretrained parking-slot detector is then applied to both IPM images to extract task-relevant structural features. In this paper, we adopt two widely used methods, DMPR-PS~\cite{huang2019dmpr} and GCN-Parking~\cite{min2021attentional_gcn}, finetuned on ParkRecon3D. During reconstruction optimization, the perception network is kept frozen, providing stable, non-drifting guidance. 
\subsubsection{Parking Corner Guidance}
Taking DMPR-PS as an example, we feed the ground-truth and rendered IPM images through the same network:
\begin{equation}
H^{\text{t}} = f_{\text{DMPR-PS}}\!\left(I_{\text{ipm}}^{\text{gt}}\right), 
\quad
H^{\text{s}} = f_{\text{DMPR-PS}}\!\left(\hat{I}_{\text{ipm}}\right)
\end{equation}
where $H^{\text{t}}$ and $H^{\text{s}}$ denote the teacher and student feature maps, respectively, each containing corner confidence, direction fields, and offset predictions.  
Since our goal is to make the structural features extracted from the rendered IPM $\hat{I}_{\text{ipm}}$ approach those extracted from the ground-truth IPM $I_{\text{ipm}}^{\text{gt}}$, we treat $H^{\text{t}}$ as the teacher signal and $H^{\text{s}}$ as the student representation. The corner-confidence channels of $H^{\text{t}}$ and $H^{\text{s}}$ are then used to construct slot-aware weight maps, guiding the reconstruction toward slot-critical geometric structures. 
 
Let $H^{\text{t}}_{\text{conf}}$ and $H^{\text{s}}_{\text{conf}}$ denote the slot corner confidence channels of $H^{\text{t}}$ and $H^{\text{s}}$, respectively.  
We convert these confidence maps into continuous slot-aware weights by applying a differentiable shaping function: 
\begin{equation}
W = \sigma\!\left(\frac{H_{\text{conf}} - \tau}{T}\right)^{\gamma}
\end{equation}
where $T=0.5$ is a temperature parameter controlling the softness of the mask, 
$\tau=0.25$ is a confidence threshold, and $\gamma=1$ produces a linear shaping. 
These choices follow our implementation and empirically balance sharp localization with stable gradients. 
This transformation yields soft masks $W^{\text{t}}$ and $W^{\text{s}}$ that concentrate supervision around high-confidence slot corners.

Based on the soft masks $W^{\text{t}}$ and $W^{\text{s}}$, we construct a
mixed slot-aware weight map that balances reliable geometric supervision
from ground truth detections with adaptive cues from the rendered predictions:
\begin{equation}
W_{\text{mix}}
= \alpha\, W^{\text{t}}
+ (1-\alpha)\, \mathrm{sg}\!\left(W^{\text{s}}\right),
\quad
\alpha = 0.8
\end{equation}

This formulation serves two purposes.  
First, the stop-gradient operator $\mathrm{sg}(\cdot)$ prevents the student
weights from being directly updated, avoiding degenerate solutions such as
uniformly low confidence and ensuring that gradients flow only through the
reconstruction pathway.  
Second, the mixture combines the stability of teacher-derived supervision
with the adaptiveness of student predictions, allowing the model to focus on
slot-critical geometric regions that may vary across frames. 

The resulting mixed weights $W_{\text{mix}}$ are upsampled to the
resolution of the rendered IPM image and subsequently back-projected into
each surround-view fisheye camera.  
This ensures that slot-aware supervision is consistently applied both in
the IPM space and the raw fisheye camera space. 
\subsubsection{Slot Edge-aware Extension}
GCN-Parking~\cite{min2021attentional_gcn} extends DMPR-PS by additionally
predicting slot edges between detected corners.  
Let $A^{\mathrm{t}}$ and $A^{\mathrm{s}}$ denote the teacher and student
edge-score matrices obtained from the ground-truth and rendered IPM images,
respectively.  
For each image, we select the top-$K_e$ edge candidates
($K_e = 8$ in our experiments) and, for every edge $(i,j)$, define a line
segment between the corresponding corner locations $p_i$ and $p_j$:
\begin{equation}
\ell_{ij}(t) = (1-t)\,p_i + t\,p_j, \quad t\in[0,1]
\end{equation}
We then rasterize each edge onto the IPM grid using a Gaussian tube with
width $\sigma = 1.5$ and $N_s = 32$ samples:
\begin{equation}
e_{ij}(u,v)
=
\max_{t\in[0,1]}
\exp\!\left(
-\frac{\lVert (u,v)-\ell_{ij}(t)\rVert_2^2}{2\sigma^2}
\right)
\end{equation}
Aggregating all selected edges $e_{ij}(u,v)$ yields teacher and student edge weight maps
$W_{\text{edge}}^{\mathrm{t}}$ and $W_{\text{edge}}^{\mathrm{s}}$, which are
mixed analogously to the corner masks,
\begin{equation}
W_{\text{edge}}
=
\beta\, W_{\text{edge}}^{\mathrm{t}}
+
(1-\beta)\,\mathrm{sg}\!\left(W_{\text{edge}}^{\text{s}}\right),
\quad
\beta = 0.8
\end{equation}
and combined with the DMPR-PS–based mask to form the final slot-aware
weights:
\begin{equation}
W_{\text{mix}}'
=
W_{\text{mix}}
+
\lambda_{\text{edge}}\, W_{\text{edge}}
\end{equation}
which emphasizes both corner points and slot boundaries during reconstruction.

\subsection{Training}

During optimization, ParkGaussian is first supervised by photometric rendering loss in vanilla 3DGS for 20,000 iterations, then optimized with alignment loss and slot aware loss for 10,000 iterations. 

\paragraph{Photometric Rendering Loss.}
Given rendered fisheye views $\hat{I}_\text{sur}$ and ground-truth images $I_\text{sur}$, we use the standard 3DGS photometric objective:
\begin{equation}
\mathcal{L}_{\mathrm{rgb}}
=
(1-\lambda)
\lVert \hat{I}_\text{sur}-I_\text{sur} \rVert_1
+
\lambda\,
\mathcal{L}_{\mathrm{D\text{-}SSIM}}
\end{equation}
where $\lambda = 0.2$.

\paragraph{Alignment Loss.}
Rendered IPM maps and ground-truth IPM maps are passed through the slot detection model to
produce student and teacher confidence fields $(W^{\mathrm{s}}, W^{\mathrm{t}})$. 
We further regularize the teacher and student weights to have similar
spatial distributions on the teacher’s top-$K$ region.  
Let $\Omega = \operatorname{TopK}(W^{\mathrm{t}})$ and define the normalized
distributions
\[
\pi^{\mathrm{s}} = \operatorname{softmax}\!\left(W^{\mathrm{s}} \mid \Omega\right),
\quad
\pi^{\mathrm{t}} = \operatorname{softmax}\!\left(W^{\mathrm{t}} \mid \Omega\right)
\]
The alignment loss is then calculated by: 
\begin{equation}
\mathcal{L}_{\mathrm{align}}
=
\mathrm{KL}\!\left(
\pi^{\mathrm{s}} \,\|\, \pi^{\mathrm{t}}
\right)
\end{equation}

\paragraph{Slot-aware reconstruction loss.}
The mixed structural weights $W_{\mathrm{mix}}$ supervise reconstruction 
both in the IPM domain and in the camera domain.  
First, weighted IPM consistency encourages the rendered IPM map 
$\hat{I}_{\mathrm{ipm}}$ to match the ground-truth $I_{\text{ipm}}^{\text{gt}}$:
\begin{equation}
\mathcal{L}_{\mathrm{ipm}}
=
\frac{
\left\lVert
W_{\mathrm{mix}}
\odot
\lvert \hat{I}_{\mathrm{ipm}} - I_{\text{ipm}}^{\text{gt}}\rvert
\right\rVert_1
}{
\lVert W_{\mathrm{mix}}\rVert_1
}
\end{equation}
The same weights are then back-projected to each camera and modulate RGB reconstruction errors:
\begin{equation}
\mathcal{L}_{\mathrm{cam}}
=
\frac{
\left\lVert
W_{\mathrm{mix}}
\odot
\lvert \hat{I}_\text{sur} - I_\text{sur} \rvert
\right\rVert_1
}{
\lVert W_{\mathrm{mix}}\rVert_1
}
\end{equation}
Together, these two terms form the slot-aware reconstruction objective that 
emphasizes geometry aligned with parking-slot structures. 

\paragraph{Final Loss.}
The total loss used during ParkGaussian optimization is:
\begin{equation}
\mathcal{L}
=
\mathcal{L}_{\mathrm{rgb}}
+
\lambda_{\mathrm{align}}\,\mathcal{L}_{\mathrm{align}}
+
\lambda_{\mathrm{ipm}}\,\mathcal{L}_{\mathrm{ipm}}
+
\lambda_{\mathrm{cam}}\,\mathcal{L}_{\mathrm{cam}}
\end{equation}
where $\lambda_\mathrm{align}=0.001$, $\lambda_{\mathrm{ipm}}=0.1$, $\lambda_{\mathrm{cam}}=0.1$.
\section{Experiments}
\subsection{Experimental Setup}

\paragraph{Baselines.}
For reconstruction, we restrict comparison to a method explicitly validated for parking environments.
Following previous protocols, we adopt OmniRe~\cite{chen2024omnire}, 3DGUT~\cite{qiwu3dgut}, and Self-Cali-GS~\cite{selfcalibgs2025} that mainly focus on fisheye camera environments as a reconstruction baseline.
For parking slot detection, we select the surround-view systems method~\cite{huang2019dmpr, min2021attentional_gcn} as a baseline, which differs from monocular camera-based methods.
Furthermore, we fine-tune them to better align with our benchmark.
During the training, we use four scenes of our benchmark, from which 100 frames of four‑channel surround fisheye inputs are sampled for each scene. Evaluation is performed every 10 frames to compute the relevant metrics.
For the perception models, we use the corresponding frames’ ground‑truth IPM images to train the two detection networks.
\textbf{Due to the space limitation, please refer to the appendix for more experiments. }

\paragraph{Evaluation Metrics.} We test the underground garage scene on Self-Cali-GS~\cite{selfcalibgs2025}, OmniRe~\cite{chen2024omnire}, 3DGUT~\cite{qiwu3dgut} and our method, and compute the Peak Signal-to-Noise Ratio (PSNR), Structural Similarity Index Measure (SSIM), and Learned Perceptual Image Patch Similarity (LPIPS) metrics for the entire image to evaluate the impact of the underground garage's challenged conditions on 3D reconstruction. For the downstream task of parking slot detection, we use DMPR-PS and GCN Parking Slot methods to calculate Precision and Recall to evaluate the detection metrics of the reconstructed scene from a synthetic BEV perspective. The parking point Precision and Recall are defined as:
\begin{equation}
\text{precision} = \frac{\text{true positives}}{\text{true positives} + \text{false positives}}
\end{equation}

\begin{equation}
\text{recall} = \frac{\text{true positives}}{\text{true positives} + \text{false negatives}}
\end{equation}
Suppose the marked real parking space entry points are \( P_{r_1}(x_{r_1},y_{r_1}), P_{r_2}(x_{r_2},y_{r_2})\), and the identified parking space entry points are \( P_{i_1}(x_{i_1},y_{i_1}), P_{i_2}(x_{i_2},y_{i_2})\). If the relationship between two points satisfies:
\begin{equation}
\|(p_{r_1} - p_{i_1}, p_{r_2} - p_{r_2})\|_2 < distance
\end{equation}
And the confidence meets its threshold, then it will be recognized as a true positive; otherwise, it is a false positive. If none of the above conditions are met, it will be recognized as a false negative. Where the \(distance \) can be adjusted. On this basis, the parking bitmap angle constructed from the parking entrance point is introduced, and the Precision and the Recall of the predicted parking space can be calculated by judging whether it meets: \( |{\theta}_r - {\theta}_i| < angle\) or \(360^\circ - |{\theta}_r - {\theta}_i| < angle\), where the \(angle \) can be adjusted.

\paragraph{Implementation Details.} 
Our ParkGaussian framework is implemented in PyTorch and trained for 30{,}000
iterations on a single NVIDIA RTX~4090 GPU using the Adam optimizer. For 3D
Gaussian attributes, we adopt the learning-rate schedule used in 3DGS~\cite{2023_8_08-3dgs_for_real_time_radiance_field_rendering} and 3DGUT~\cite{qiwu3dgut}, where the position learning rate follows an exponential decay to 1\% of its initial value. We also employ the Markov Chain Monte Carlo optimization strategy~\cite{kheradmand20243d} implemented in GSplat~\cite{ye2025gsplat} to improve convergence and reconstruction stability. 
The Gaussian primitives are initialized from the COLMAP sparse point cloud in ParkRecon3D. 
All remaining hyperparameters follow those described in the method section.

\subsection{Main Results}
\subsubsection{Novel View Synthesis} 

\begin{table}[htbp]
 \footnotesize
  \centering
  \caption{Novel view synthesis performance on the proposed ParkRecon3D dataset. \textbf{Bold} denotes the best result and \underline{underline} denotes the second best.}  
  \label{tab:reconstruction_performance} 
  \begin{tabular}{ccccc}
    \toprule
    \textbf{Scene} & \textbf{Modules}   & \textbf{PSNR}$\uparrow$ & \textbf{SSIM}$\uparrow$          & \textbf{LPIPS}$\downarrow$ \\
    \midrule
    \multirow{5}{*}{Scene1}
        & Self-Cali-GS~\cite{selfcalibgs2025} & 23.78  & 0.82   & 0.31 \\
        & 3DGUT~\cite{qiwu3dgut}              & 28.70  & \underline{0.92} & \underline{0.21} \\
        & OmniRe~\cite{chen2024omnire}        & 25.12  & 0.84   & 0.37 \\
        & Ours (w/ DMPR-PS)                   & \underline{29.10} & \underline{0.92} & \textbf{0.20} \\
        & Ours (w/ GCN-Parking)               & \textbf{30.09} & \textbf{0.93} & \textbf{0.20} \\
    \midrule
    \multirow{5}{*}{Scene3}
        & Self-Cali-GS~\cite{selfcalibgs2025} & 20.10 & 0.81 & 0.30 \\
        & 3DGUT~\cite{qiwu3dgut}              & 27.80 & \underline{0.92} & \textbf{0.20} \\
        & OmniRe~\cite{chen2024omnire}        & 21.58 & 0.78 & 0.50 \\
        & Ours (w/ DMPR-PS)                   & \underline{28.89} & \underline{0.92} & \underline{0.21} \\
        & Ours (w/ GCN-Parking)               & \textbf{30.27} & \textbf{0.93} & \textbf{0.20} \\
    \bottomrule
  \end{tabular}
\end{table}

\begin{figure*}[htbp] 
  \centering
  \includegraphics[width=1.0\textwidth]{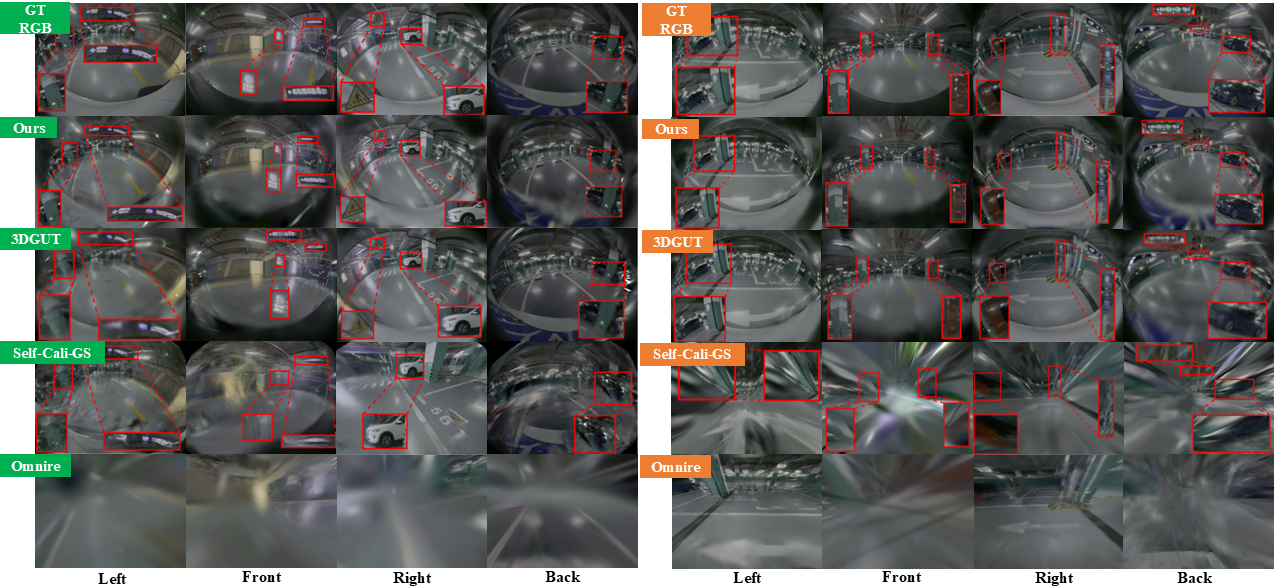} 
  \caption{Novel view synthesis visualization in 4 surround view fisheye images.}
  \label{fig:experiment_compare.png} 
  \vspace{-4mm}
\end{figure*}

Tab.~\ref{tab:reconstruction_performance} shows the results on the proposed ParkRecon3D dataset. The results show that ParkingGaussian achieves state-of-the-art performance in the Benchmark.
The previous method has a poor performance because they were merely emphasizes visual fidelity while overlooking the fundamental goal of simulation.
For a more intuitive understanding of the model's reconstruction performance, we present the side‑by‑side visual comparison between the previous work and ours in Fig.~\ref{fig:experiment_compare.png}, which evident that under the inherent constraints of underground parking garages, although 3DGUT~\cite{qiwu3dgut} and Self-Cali-GS~\cite{selfcalibgs2025} can construct the spatial topology of the overall scene, they have obvious limitations in the robustness of detail representation. OmniRe~\cite{chen2024omnire} exhibits significantly inferior reconstruction quality, with severe blurring and structural loss across all perspectives. 

Compared with all these methods, our approach achieves high‑level visual alignment with the ground‑truth scenes from all perspectives and enables more accurate scene reconstruction by incorporating a slot‑aware reconstruction strategy that leverages parking‑slot detectors to enhance reconstruction fidelity in task‑critical slot regions. 

\newpage 
\subsubsection{Parking Slot Detection}

\begin{wrapfigure}{r}{0.5\columnwidth} 
  \centering
  \includegraphics[width=\linewidth]{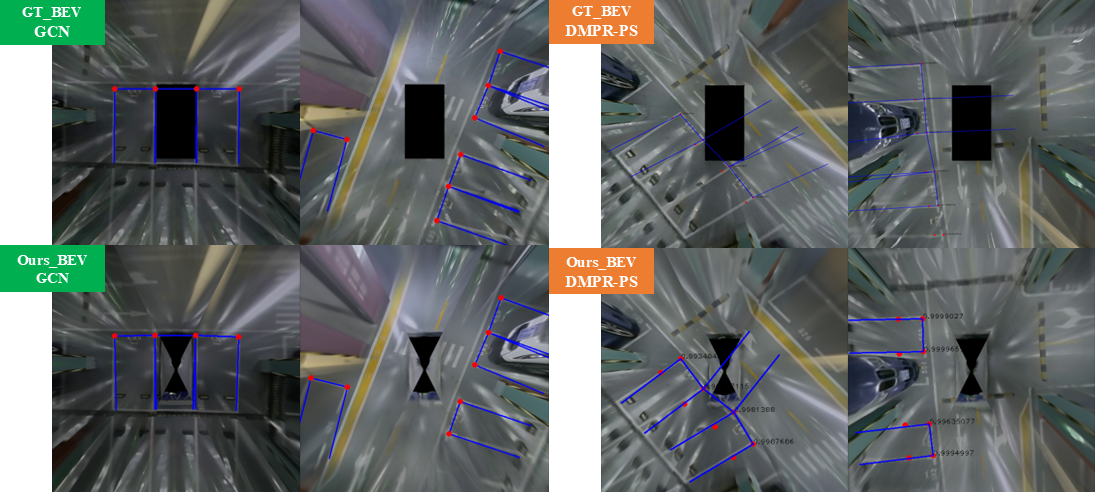}
  \caption{Parking slot detection visualization in IPM images.}
  \label{fig:plot_compare.png}
  \vspace{-10pt} 
\end{wrapfigure}

Tab.~\ref{tab:perception_performance1},~\ref{tab:perception_performance2} demonstrate that our slot‑aware reconstruction strategy substantially enhances the performance of parking slot detection.
Both perception networks achieve performance close to the ground truth when operating on our reconstruction model, and both show noticeable improvements after incorporating the perception module during reconstruction.
In Fig~\ref{fig:plot_compare.png}, we provide a visual comparison between the detection results on real images and those obtained from the renderings of our reconstruction model. 
As can be visually observed, our approach maintains high‑fidelity reconstruction of the global scene and, at the same time, accurately reproduces the fine‑grained structures of the parking slots.

\begin{table*}[t] 
  \centering
  \begin{minipage}{0.49\textwidth} 
    \centering
    \caption{DMPR-PS model comparison: results on improved AVM-SLAM dataset for parking slot point detection.}
    \label{tab:perception_performance1}
    \resizebox{\linewidth}{!}{%
      \begin{tabular}{ccccc}
        \toprule
        \multirow{2}{*}{\textbf{Variant}} & \multicolumn{2}{c}{\textbf{Scene1}} & \multicolumn{2}{c}{\textbf{Scene3}} \\
        \cmidrule(lr){2-3} \cmidrule(lr){4-5}
        & Precision$\uparrow$ & Recall$\uparrow$ & Precision$\uparrow$ & Recall$\uparrow$ \\
        \midrule
        GT-DMPR-PS~\cite{huang2019dmpr} & 0.86 & 0.22 & 0.49 & 0.21 \\
        Ours w/o Per DMPR-PS            & 0.71 & 0.08 & 0.48 & 0.18 \\
        Ours w/ Per DMPR-PS             & 0.74 & 0.10 & 0.47 & 0.19 \\
        \bottomrule
      \end{tabular}
    }
  \end{minipage}
  \hfill 
  \begin{minipage}{0.49\textwidth} 
    \centering
    \caption{GCN model comparison: results on improved AVM-SLAM dataset for parking slot detection.}
    \label{tab:perception_performance2}
    \resizebox{\linewidth}{!}{%
      \begin{tabular}{ccccc}
        \toprule
        \multirow{2}{*}{\textbf{Variant}} & \multicolumn{2}{c}{\textbf{Scene1}} & \multicolumn{2}{c}{\textbf{Scene3}} \\
        \cmidrule(lr){2-3} \cmidrule(lr){4-5}
        & Precision$\uparrow$ & Recall$\uparrow$ & Precision$\uparrow$ & Recall$\uparrow$ \\
        \midrule
        GT-GCN~\cite{min2021attentional_gcn} & 0.99 & 0.49 & 0.98 & 0.50 \\
        Ours w/o Per GCN                     & 0.95 & 0.40 & 0.94 & 0.48 \\
        Ours w/ Per GCN                      & 0.97 & 0.43 & 0.95 & 0.48 \\
        \bottomrule
      \end{tabular}
    }
  \end{minipage}
  \vspace{-4mm} 
\end{table*}

\subsection{Ablation on Slot-aware Strategy}
We conduct comprehensive ablation studies to disentangle the contribution of each component in our slot-aware reconstruction framework. Five variants are evaluated:
(1) Direct IPM $L_{1}$ supervision without slot-aware cues;
(2) Feature-level supervision using perception outputs only; 
(3) Teacher-only weighting using ground-truth IPM features;
(4) Student-only weighting using rendered IPM features; and 
(5) The full slot-aware design that mixes teacher–student weights with distribution alignment.



\begin{table}[htbp]
\centering
\caption{
Ablation study of the proposed slot-aware reconstruction strategy on ParkRecon3D.
Metrics include rendering quality and downstream parking-slot detection.
}
\label{tab:ablation_slotaware}

\footnotesize 

\setlength{\tabcolsep}{3.5pt} 

\begin{tabular}{@{}lccccc@{}} 
\toprule
\textbf{Variant} &
\textbf{PSNR}$\uparrow$ & \textbf{SSIM}$\uparrow$ & \textbf{LPIPS}$\downarrow$ &
\textbf{Prec.}$\uparrow$ & \textbf{Rec.}$\uparrow$ \\ 
\midrule
\multicolumn{6}{c}{\textit{\textcolor{gray}{Scene1}}} \\
Direct IPM $L_1$ supervision               & 24.94 & 0.88 & 0.33 & 0.62 & 0.06 \\
Feature-level supervision only             & 27.43 & 0.91 & 0.28 & 0.64 & 0.05 \\
Teacher-only weighting                     & 29.56 & 0.92 & 0.21 & 0.90 & 0.41 \\
Student-only weighting                     & 28.62 & 0.90 & 0.22 & 0.81 & 0.40 \\
\textbf{Full slot-aware (ours)}            & \textbf{30.09} & \textbf{0.93} & \textbf{0.20} & \textbf{0.97} & \textbf{0.43} \\
\midrule
\multicolumn{6}{c}{\textit{\textcolor{gray}{Scene3}}} \\
Direct IPM $L_1$ supervision               & 24.99 & 0.89 & 0.35 & 0.58 & 0.04 \\
Feature-level supervision only             & 27.36 & 0.91 & 0.27 & 0.43 & 0.11 \\
Teacher-only weighting                     & 28.77 & 0.93 & 0.21 & 0.88 & 0.45 \\
Student-only weighting                     & 26.72 & 0.90 & 0.29 & 0.71 & 0.41 \\
\textbf{Full slot-aware (ours)}            & \textbf{30.27} & \textbf{0.93} & \textbf{0.20} & \textbf{0.95} & \textbf{0.48} \\
\bottomrule
\end{tabular}

\end{table}

The ablation results in Tab.~\ref{tab:ablation_slotaware} reveal several important insights into the effectiveness of the proposed slot-aware strategy. 
First, naively applying an IPM-space $L_{1}$ loss yields poor reconstruction quality and extremely weak parking-slot detection performance. This is largely because multi-view projections introduce conflicts at view boundaries, injecting additional noise into the IPM domain.
Feature-level supervision improves rendering scores but still fails to recover reliable slot geometry, indicating that the optimization objectives of the perception model and the reconstruction model are not aligned. Their feature distributions differ substantially, causing the reconstructed geometry to deviate from the structures required by the perception network. 
Both teacher-only and student-only weighting substantially boost performance by focusing supervision on slot-relevant regions. 
However, they exhibit complementary behaviours: 
teacher-only weighting yields stable but less adaptive supervision, while student-only weighting adapts to rendered predictions but is more susceptible to noise. 
Our full slot-aware strategy, which blends teacher–student weighting and imposes distribution alignment, consistently achieves the best results across both scenes. 
It achieves the highest rendering quality and, by aligning reconstruction with the perception model, significantly improves downstream precision and recall, showing that integrating structural priors with prediction consistency is essential for robust parking-aware reconstruction.

\section{Limitation}
\begin{figure}[!ht]
  \centering
  \begin{subfigure}[b]{0.15\textwidth}
    \centering
    \includegraphics[width=\linewidth]{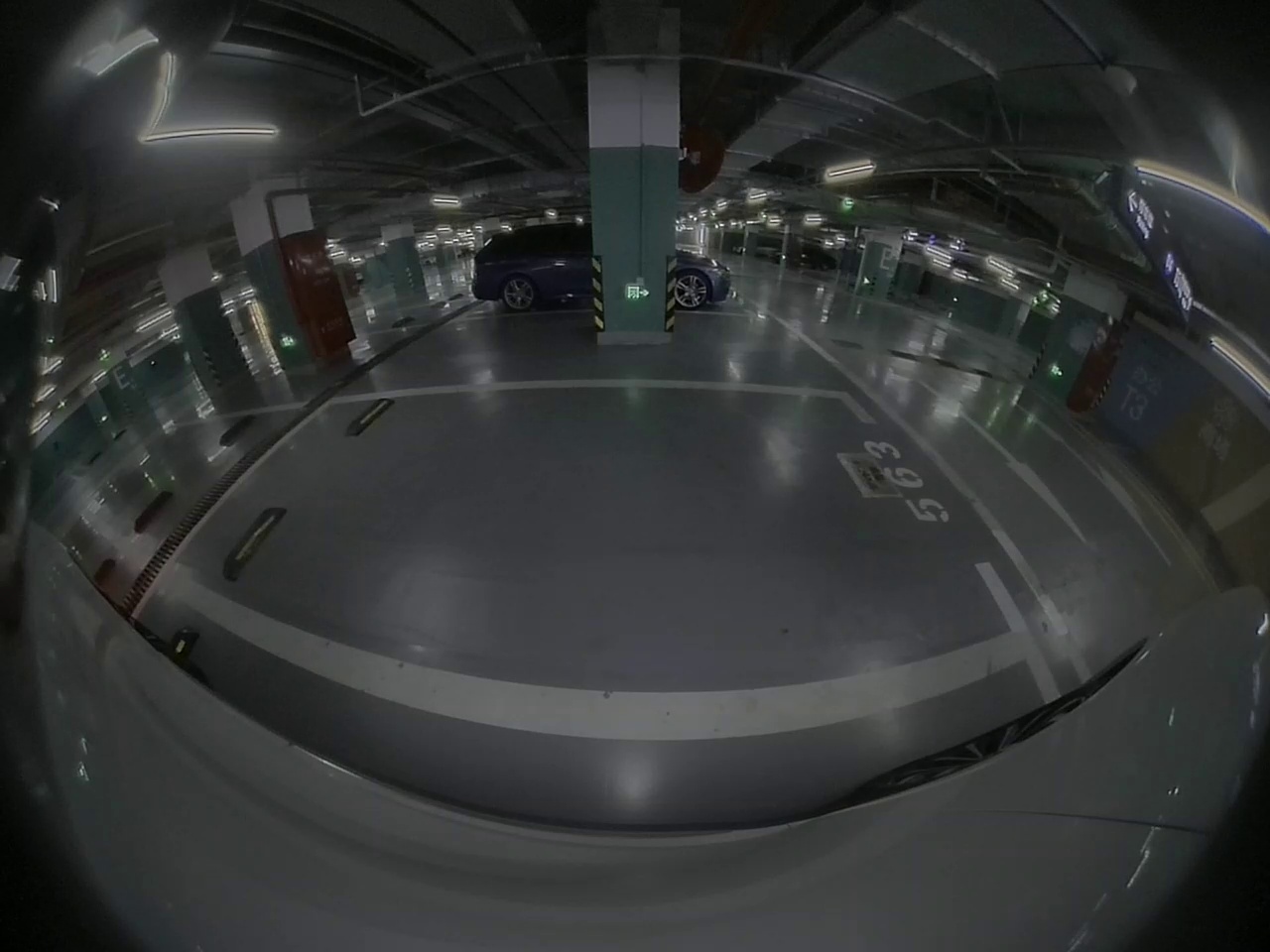}  
    \caption{Reflections}
  \end{subfigure}
  \hspace{2mm} 
  \begin{subfigure}[b]{0.15\textwidth}
    \centering
    \includegraphics[width=\linewidth]{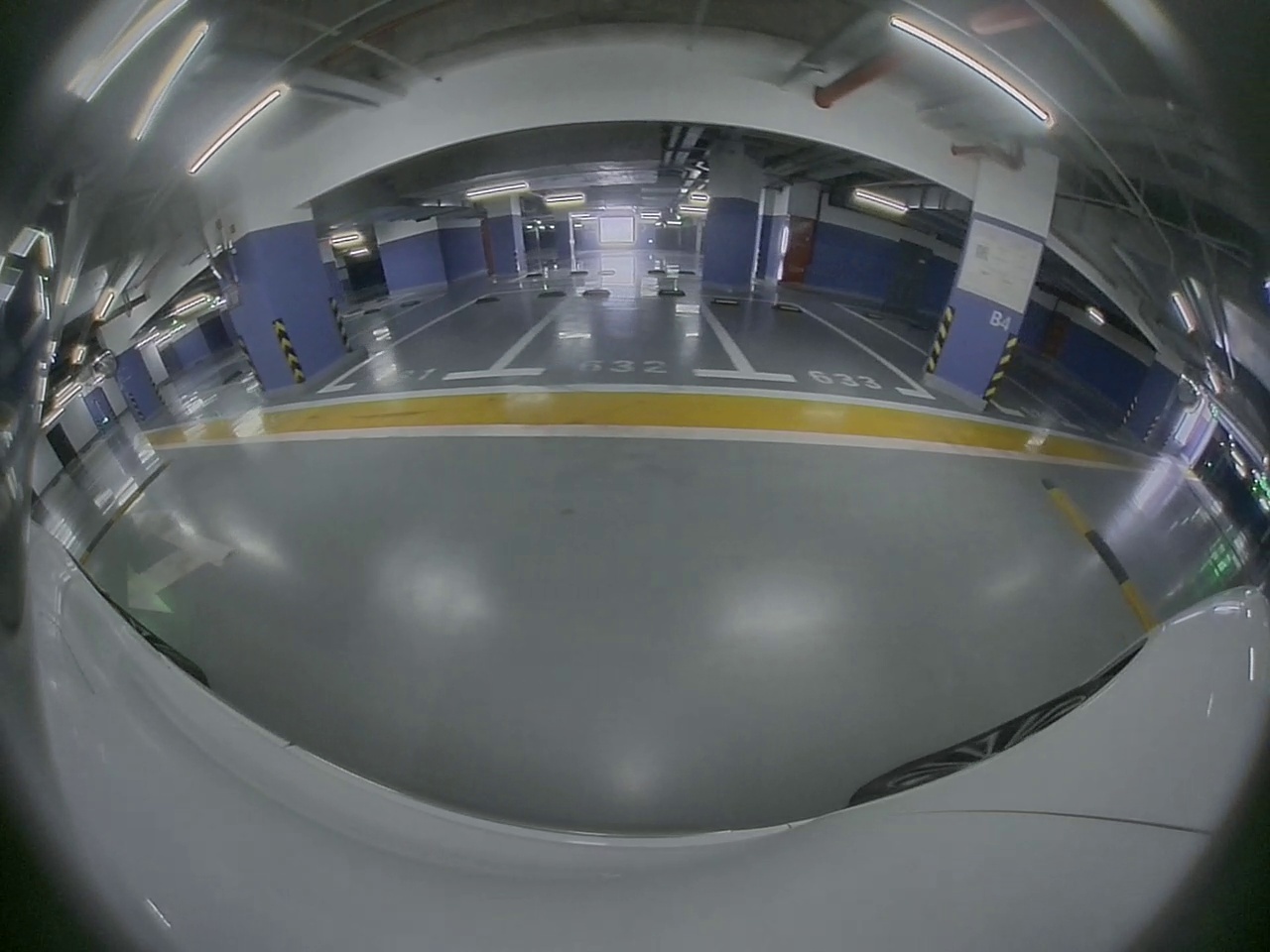}  
    \caption{Textures}
  \end{subfigure}
  \hspace{2mm} 
  \begin{subfigure}[b]{0.15\textwidth}
    \centering
    \includegraphics[width=\linewidth]{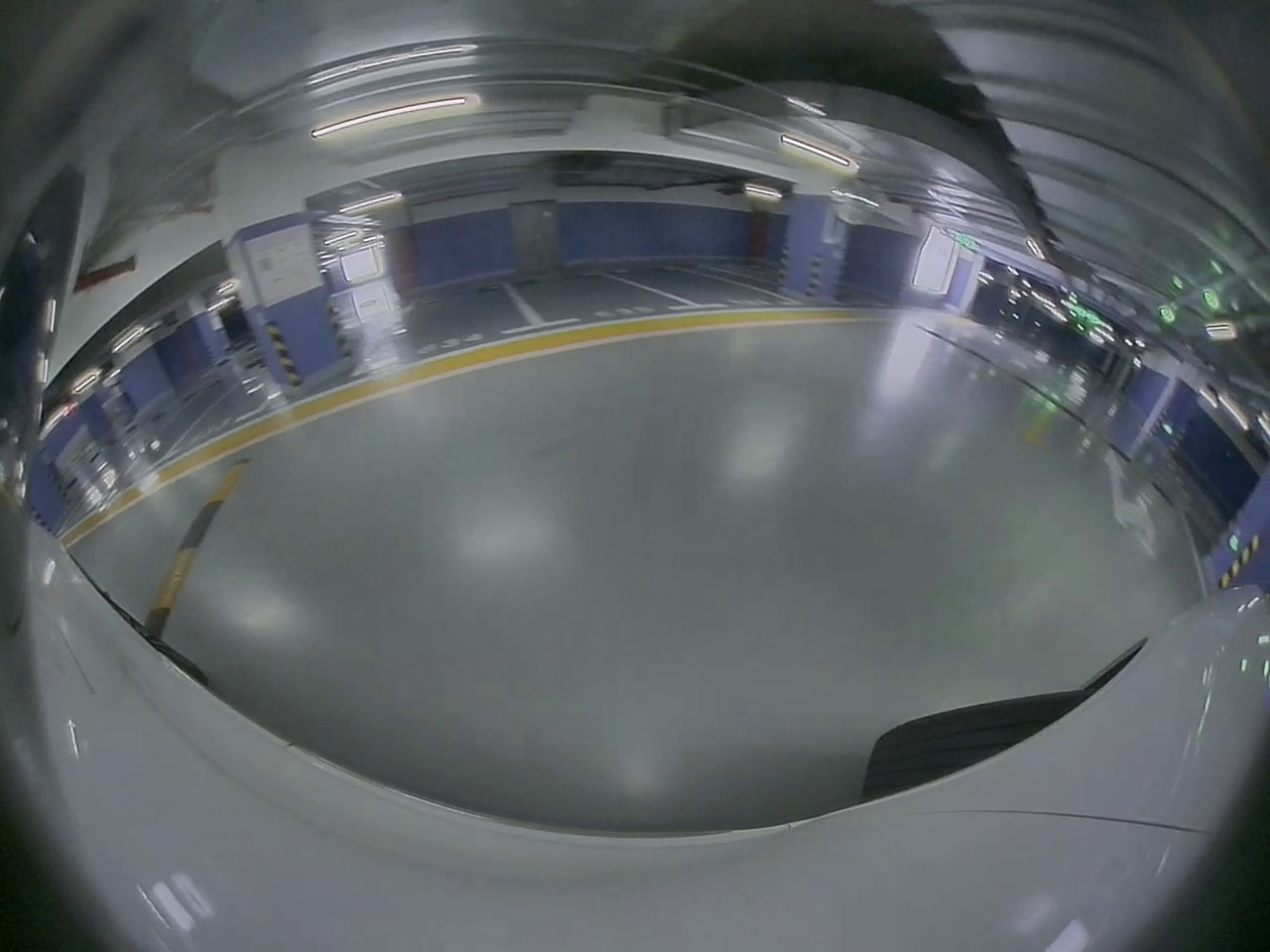}
    \caption{Blur}
  \end{subfigure}
  
  \vspace{-2mm}
  \caption{Challenges in underground parking lots.}
  \vspace{-4mm}
  \label{fig:parking_difficulties}
\end{figure}

ParkRecon3D still contains several inherent challenges of underground parking environments, including indoor specular reflections, highly repetitive textures, and motion blur caused by long exposure under weak lighting (Fig.~\ref{fig:parking_difficulties}). These characteristics are difficult to model accurately and will be further addressed in future work. 

\section{Conclusion}
In this work, we present ParkRecon3D, a novel framework that integrates 3D Gaussian Splatting with a slot-aware reconstruction strategy for high-quality parking scene modeling. Built upon the proposed ParkRecon3D benchmark, our method achieves accurate and efficient 3D reconstruction in underground parking environments while enhancing synthesis quality in parking slot regions. ParkRecon3D provides a reliable foundation for the development and evaluation of future autonomous parking systems.



{
    \small
    \bibliographystyle{ieeenat_fullname}
    \bibliography{paper}
}

\end{document}